\RequirePackage{fix-cm}
\documentclass[smallextended]{svjour3}       
\smartqed  
\usepackage{amsmath,amsfonts,amssymb}
\usepackage{graphicx}
\usepackage{algorithm}
\usepackage[noend]{algpseudocode}
\usepackage{array}
\usepackage{hyperref}
\graphicspath{{./Figures/}} 

\renewcommand{\arraystretch}{2.5}

\begin{document}

\title{A Deep Learning Spatiotemporal Prediction Framework for Mobile Crowdsourced Services
}
\author{Ahmed Ben Said         \and
        Abdelkarim Erradi \and
        Azadeh Ghari Neiat \and
        Athman Bouguettaya
}


\institute{Ahmed Ben Said \at
              Department of Computer Science and Engineering, College of Engineering \\
              Qatar University, Doha, Qatar\\
              \email{abensaid@qu.edu.qa}           
           \and
           Abdelkarim Erradi \at
			Department of Computer Science and Engineering, College of Engineering \\
			Qatar University, Doha, Qatar\\
			\email{erradi@qu.edu.qa}    
			\and
			Azadeh Ghari Neiat \at
			School of Information Technologies, University of Sydney \\
			Australia\\
			\email{azadeh.gharineiat@sydney.edu.au}  
			\and
			Athman Bouguettaya \at
			School of Information Technologies, University of Sydney \\
			Australia\\
			\email{athman.bouguettayat@sydney.edu.au}  			       
}

\date{Received: date / Accepted: date}

\authorrunning{Ben Said et al.}
\titlerunning{ }
\maketitle

\begin{abstract}
This papers presents a deep learning-based framework to predict crowdsourced service availability spatially and temporally. A novel two-stage prediction model is introduced based on historical spatio-temporal traces of mobile crowdsourced services. The prediction model first clusters mobile crowdsourced services into regions. The availability prediction of a mobile crowdsourced service at a certain location and time is then formulated as a classification problem. To determine the availability duration of predicted mobile crowdsourced services, we formulate a  forecasting task of time series using the Gramian Angular Field. We validated the effectiveness of the proposed framework through multiple experiments.  

\end{abstract}


\section{Introduction}
\label{intro}
Thanks to the fast proliferation of mobile devices, equipped with increasingly powerful sensing capabilities, Mobile Crowdsourcing (MCS) has emerged as a new paradigm that allows obtaining required data and services by soliciting contribution from the crowd. Storing, processing and handling crowdsourced sensory data also inherit important challenges. Cloud services could be used to ease storing, aggregating and managing the spatiotemporal data collected from MCS. By harnessing the service paradigm, the crowdsourced sensor data can be abstracted as cloud-hosted crowdsourced services \cite{ref1,ref2}.\newline
Mobile crowdsourced services can act as consumers and providers (e.g., WiFi hotspots). In this work, we focus on crowdsourced WiFi hotspots. An intrinsic feature of WiFi Hotspots services is their mobility. It provides great opportunities but also challenges. One such a challenge is the ability to predict the location and the availability duration of crowdsourced services.  The key aspects of crowdsourced hotspots are the spatio-temporal attributes. They are central to the selection and composition of WiFi hotspot services. In our previous work \cite{ref1,ref2}, crowdsourced services are considered to be deterministic, i.e., the service attributes (functional and non-functional) are known a priori. However, such assumption only allows static queries for services with known location and availability time. This paper considers non-deterministic crowdsourced services whose time and availability are unknown in advance. For Web services, Zeng et al. \cite{ref3} define availability as the probability of a service being accessible. Silic et al. \cite{ref4} define availability as the probability that a service invocation will complete successfully under the specified time constraints. From a crowdsourced service perspective, the availability is constrained by the spatio-temporal attributes of the service itself as the crowdsourced service is available only at a given location for a particular time slot.   \newline
This paper presents a deep learning-based strategy to determine if a crowdsourced service (e.g., WiFi hotspot) is available given particular location and timing. Our objective is to address the following challenges: Which crowdsourced service is available? When and where will this service(s) be available? By acquiring the knowledge about the spatio-temporal attributes, services can then be composed with greater certainty.  
In summary, this work is informed by the following facts:
\begin{itemize}
	\item Sensor data, particularly the ones from smartphones, can be collected and acquired easily. The abundance of these data offers a great opportunity to leverage useful information related to users such as their daily activities and routines, geo-spatial and temporal behavior, etc. Therefore, one can infer, given the appropriate tools and techniques, the availability of someone in the spatio-temporal domain. 
	\item 
	Deep learning has been widely applied in computer vision, intelligent transportation system \cite{rj} healthcare \cite{bens} and many other applications including spatiotemporal data collected by smartphones. These advances have been guided by steady supply of data. 
\end{itemize}
Our contributions consist of  proposing a deep learning-based framework for service availability prediction. In the first stage, each geolocation is assigned to the crowdsourced service provider(s) at particular time granularity. By using the historical data, our deep learning approach can predict available crowdsourced services along a specific region at a particular time. Once the crowdsourced available services are predicted, their availability duration can be determined in the second stage using a Gramian Angular Field time series prediction model that makes use of the historical temporal attributes of the service. \newline
The rest of the paper is as follows:  We detail the crowdsourced service model in section II. Section III presents details of the deep learning approach. In section IV, experimental results are detailed and we conclude in the last section. 

\section{Crowdsourced WiFi coverage service}
\subsection{Motivating scenario}
\begin{figure}[!h]
	\centering
	\includegraphics[scale=.5]{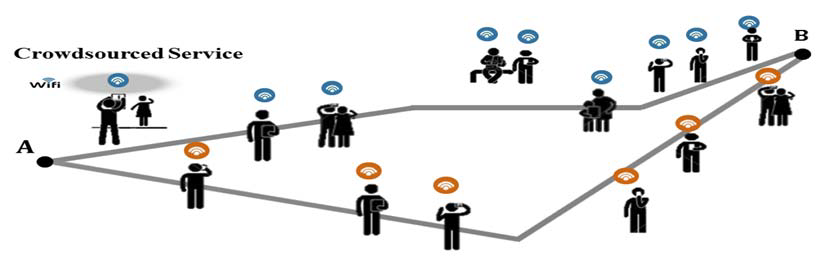}
	\caption{Crowdsourced WiFi hotspot coverage}
	\label{scenario}
\end{figure}

Fig. \ref{scenario} illustrates a WiFi crowdsourced sharing service via smartphones. For example, John may have some unused mobile data balance that he can share via a mobile application such as WiFiMapper \footnote{https://www.wifimapper.com} at a certain location for a period of time. These spatio-temporal attributes are the key to crowdsourcing hotspots to enable consumers to select suitable WiFi services. The available WiFi crowdsourced services can be overlaid on a transport network map to allow the journey planning service to recommend to users journey plans that offer the best WiFi coverage white traveling from point A to point B. In this scenario, we assume that the time and availability of  WiFi hotspot services are not known in advance.  
The problem of hotspot service prediction can be formulated as \textit{finding available crowdsourced WiFi hotspot services within a particular region at a given time based on historical spatiotemporal data of the service}.
We first present details of the spatiotemporal crowdsourced service model then we describe the proposed model to predict their spatiotemporal properties. We use crowdsourced WiFi hotspot sharing as an illustrative example.

\subsection{Crowdsourced service model }
\begin{figure}[!h]
	\centering
	\includegraphics[scale=.26]{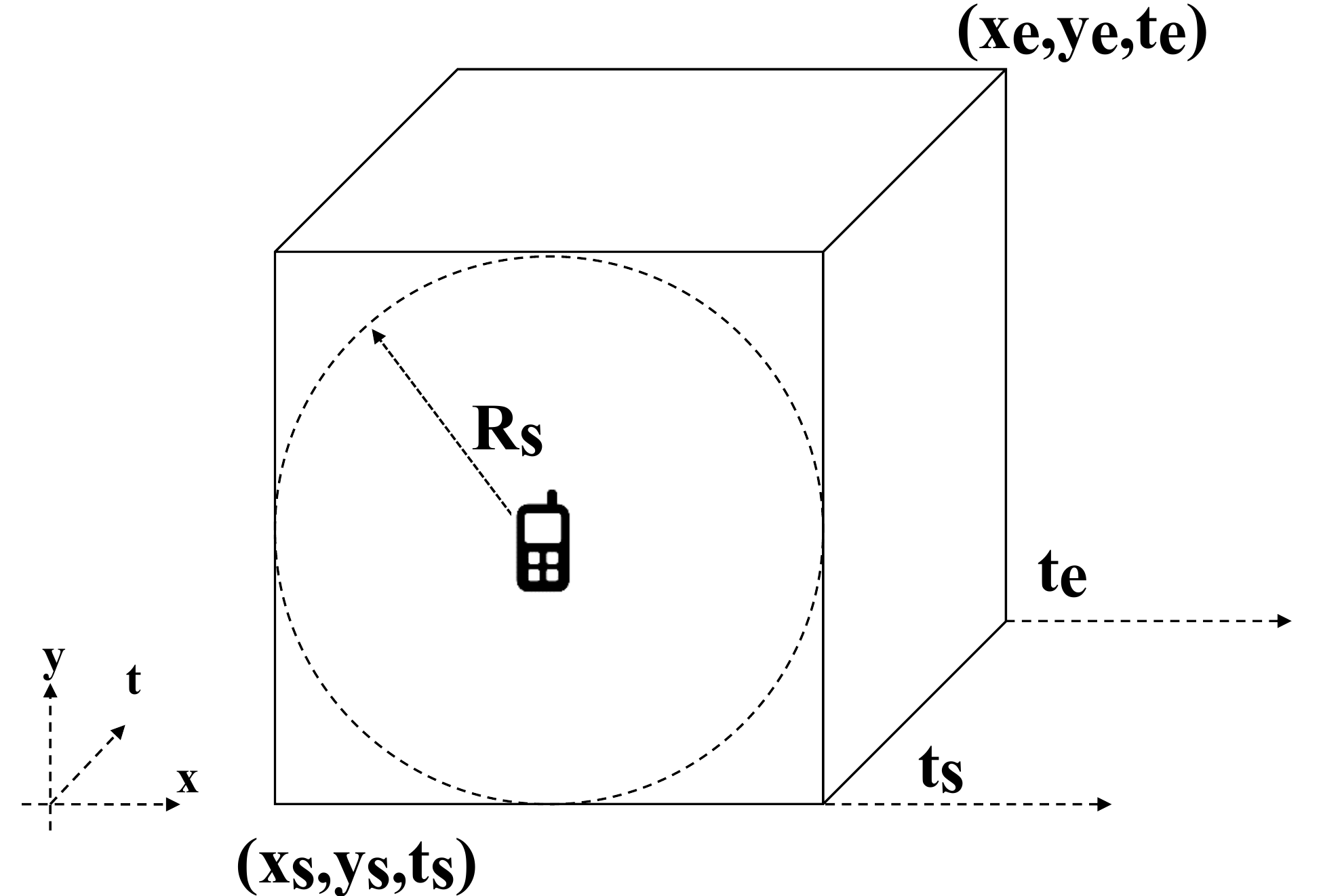}
	\caption{Crowdsourced coverage service}
	\label{model}
\end{figure}

As an illustrative example of crowdsourced service, we present in this section the WiFi coverage service model. It extends our earlier model reported in \cite{ref4,ref5} and illustrated in Fig. \ref{model}.
A sensor is a tuple \textless Sen\_id, lc, Sen\_a, tsp\textgreater where: 
\begin{itemize}
	\item Sen\_id: sensor unique id.
	\item lc: last sensor location.
	\item Sen\_a: sensing area of radius Rs whose center is loc.
	\item tsp: latest timestamps when data related to the sensor were collected. 	
\end{itemize}
A crowdsourced service S is a tuple of \textless Sid, SENS, S\&T, Fun, QoS\textgreater where: 
\begin{itemize}
	\item Sid: service unique id.
	\item SENS: finite set of sensors sen$_i$ 
	\item S\&T: space and time attributes of the service. The space is represented by the minimum bounding box of the sensing area. The time is a tuple \textless t$_s$, t$_e$\textgreater  where t$_s$ is the starting time and t$_e$ is the  end time of the service.  
	\item Fun: Functionality delivered by the service such as offering hostpot internet access. 
	\item QoS: QoS properties of the service such as bandwidth.
\end{itemize} 
Crowdsourced service can be deterministic, i.e. service attributes are known a priori, or non-deterministic where all the attributes (or subset) are not known. Discovering a service refers to finding the service id and its corresponding temporal attributes such as their availability duration. We focus our study on finding services whose spatiotemporal attributes are non-deterministic.  Investigating the non-determinism of the QoS properties of the service is left for future work. 
\newline
For a crowdsourced WiFi coverage service, the spatial characteristic is described by the WiFi coverage zone. In an outdoor scenario, the coverage extends to 100 m with perfect Line of Sight. On the other hand, the temporal attributes describing the start and end time of the service need to be predicted.

\section{Crowdsourced service availability prediction as a learning problem }
Periodicity is a common phenomenon. Animals yearly migrate to particular locations at the same particular time. Humans often have periodic activities. For example, on a weekday, one is very likely to be at home at 7 AM. At 10 AM, he is very likely to be at work, etc. Students usually have weekly schedule for classes. One can also have  his personal periodic activity such as going to the gym once or twice a week. With the wide availability of smartphones with GPS capabilities, valuable data can be collected, analyzed and used for ecological research, improving urban planning, market, ads targeting, etc. The spatio-temporal analysis presented in \cite{ref6} reveals that the individual mobility of mobile phone users is far from being random. Gonz\'{a}lez et al. \cite{ref7}  analyzed the trace of multiple mobile users over 6 months. Authors confirmed the presence high degree of spatio-temporal regularity. The main explanation of this observation is that mobile phone users spend most of the time in particular locations This fact is also confirmed by the findings of Song et al. \cite{ref8}. The study conducted by Cs\'{a}ji et al. \cite{ref9}  showed that 95\% frequently visited  less than 4 locations. The findings in \cite{ref10} demonstrated that commuters’ locations are highly predictable on average.  \newline
Based on these research findings, one can associate particular locations at a particular time granularity to particular service provider(s). Therefore, we can formulate the service availability prediction as a classification task where the aim is to find the associated label i.e. service provider(s) given the historical time-geolocation data of service providers. Furthermore, given the data of service providers, a good prediction model can estimate the time attributes of the crowdsourced service i.e. start and end time. 
\subsection{Two-stage prediction framework }
\begin{figure}[!h]
	\centering
	\includegraphics[scale=.7]{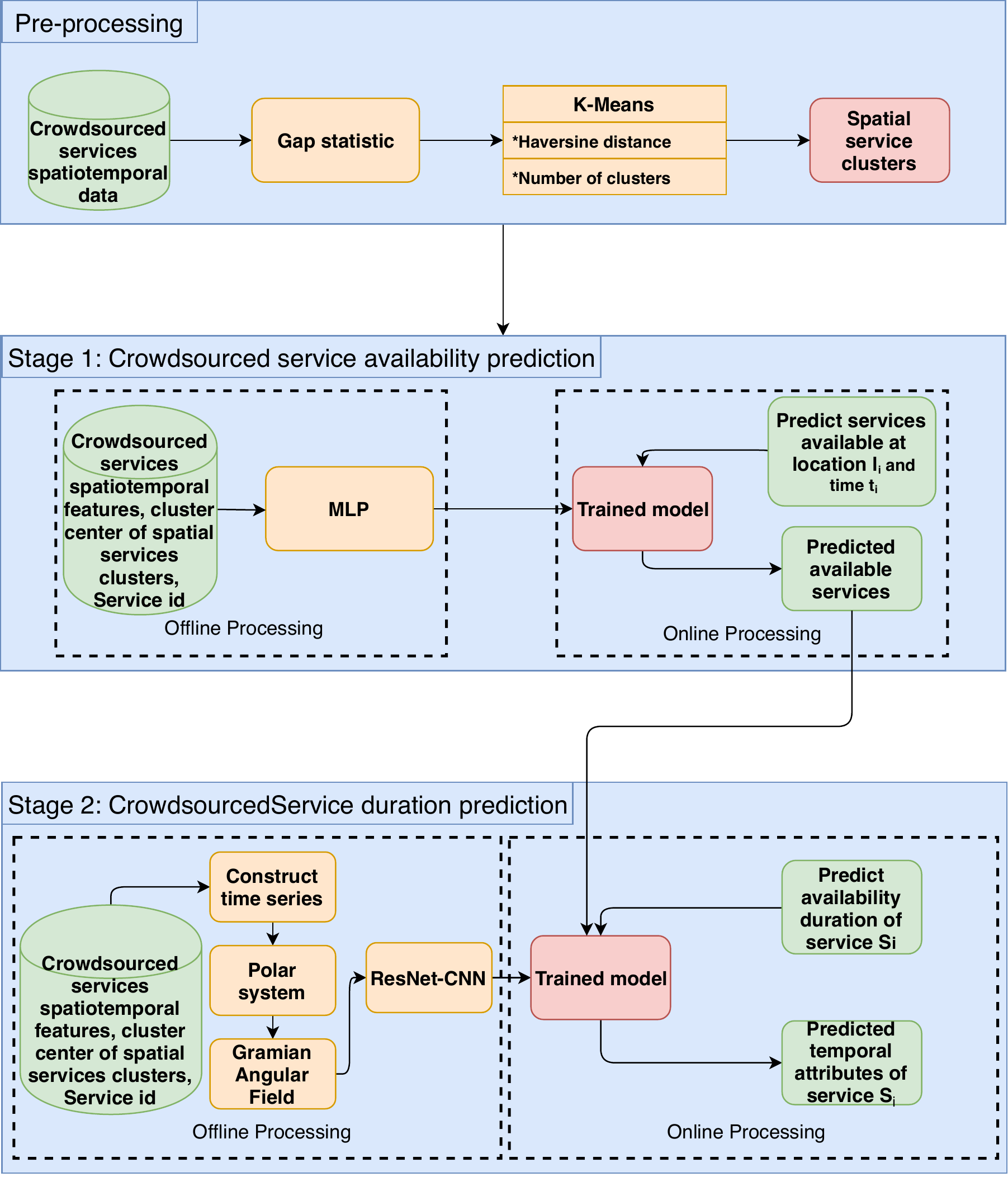}
	\caption{Two-stage prediction framework to predict crowdsourced service availability spatially and temporally}
	\label{framework}
\end{figure}

Figure \ref{framework} illustrates our proposed framework. It consists of an initial  pre-processing step of data clustering, a first stage to predict the crowdsourced service and a second stage to predict the temporal availability of the service.\newline
\noindent The initial pre-processing is a clustering task where the aim is to discover the major hotspots area where the service providers are concentrated. We propose to use  K-means algorithm \cite{ref11}. It  is probably the most well-known partitional clustering algorithm with successful application in many real-world problems such as intelligent transportation systems \cite{transport1,transport2}.
remote sensing \cite{rainfall,seviri,infrared}, semiconductor, electronics and polymer manufacturing \cite{wafer,polymer}, energy consumption and efficiency \cite{energy1,energy2,energy3}, water management \cite{water}. 
 The use of K-means is justified by its scalability, rapidity, simplicity and efficiency for processing large data \cite{big_data}. \newline Let $X = \{x_i\}$ be a set of data points where $i=1 ... N$, K-means aims at finding the set of cluster centers. This is conducted by minimizing the cost function:
\begin{equation}
\sum_{i=1}^N\sum_{x_i \in C_j}d(x_i,u_j)^2
\end{equation}
Where $C_j$ is the $j^{th}$ group or cluster of centroid $u_j$ and $d$ is a distance measure. A classic choice for $d$ is the Euclidean distance. However, unlike the classic algorithm, we propose to use the Haversine distance. It is the great-circle distance between two points. Therefore, this distance is more suitable for geolocation data. Given two geolocation points $p_1=(\phi_1,\tau_1)$ and $p_2=(\phi_2,\tau_2)$ where $\phi$ and $\tau$ are the latitude and longitude respectively, the Haversine distance $H(p_1,p_2)$ is defined as: 
\begin{equation}
    H(p_1,p_2)= R*b   
\end{equation}
Where:   
\begin{equation}
\begin{tabular}{c}
\( \phi = \phi_1 - \phi_2  \)\\
\( \tau = \tau_1 - \tau_2 \)\\
\( a = sin^2(\frac{\phi}{2})+cos(\phi_1)*cos(\phi_2)*sin(\frac{\tau}{2})   \)\\
\( b= 2*atan2(\sqrt{a},\sqrt{1-a})\)
\end{tabular}
\end{equation}
Where $R$ is the radius of the earth. Furthermore, K-means requires the number of clusters as input which is not always known a priori. To derive the number of clusters, the the gap statistic \cite{ref12} is used. Specifically, the objective is to standardize the comparison of the clustering compactness measure with a reference distribution of the data i.e. data with no straightforward clustering. The optimal number of clusters  has the lowest cluster compactness compared to the reference data. Given a cluster $C_i$ of $n_i$ points, the within-cluster dispersion is expressed as: 
\begin{equation}
	W_k =\sum_{i=1}^k\frac{1}{2n_i}D_i
\end{equation}
Where $D_i$ is the sum of pairwise distances of data points in $C_i$. The gap value is expressed as: 
\begin{equation}
Gap(k) = E^*\{log(W_k)\} - log(W_k)
\end{equation}
$E^*\{log(W_k)\}$ is calculated using Monte Carlo sampling \cite{ref13} from a reference distribution and $log(W_k)$ is derived from the sample data. Therefore, the optimal number of cluster is the one that maximizes $Gap(k)$.

\subsection{Stage 1: Crowdsourced service availability prediction}
\begin{figure}[!h]
	\centering
	\includegraphics[scale=.7]{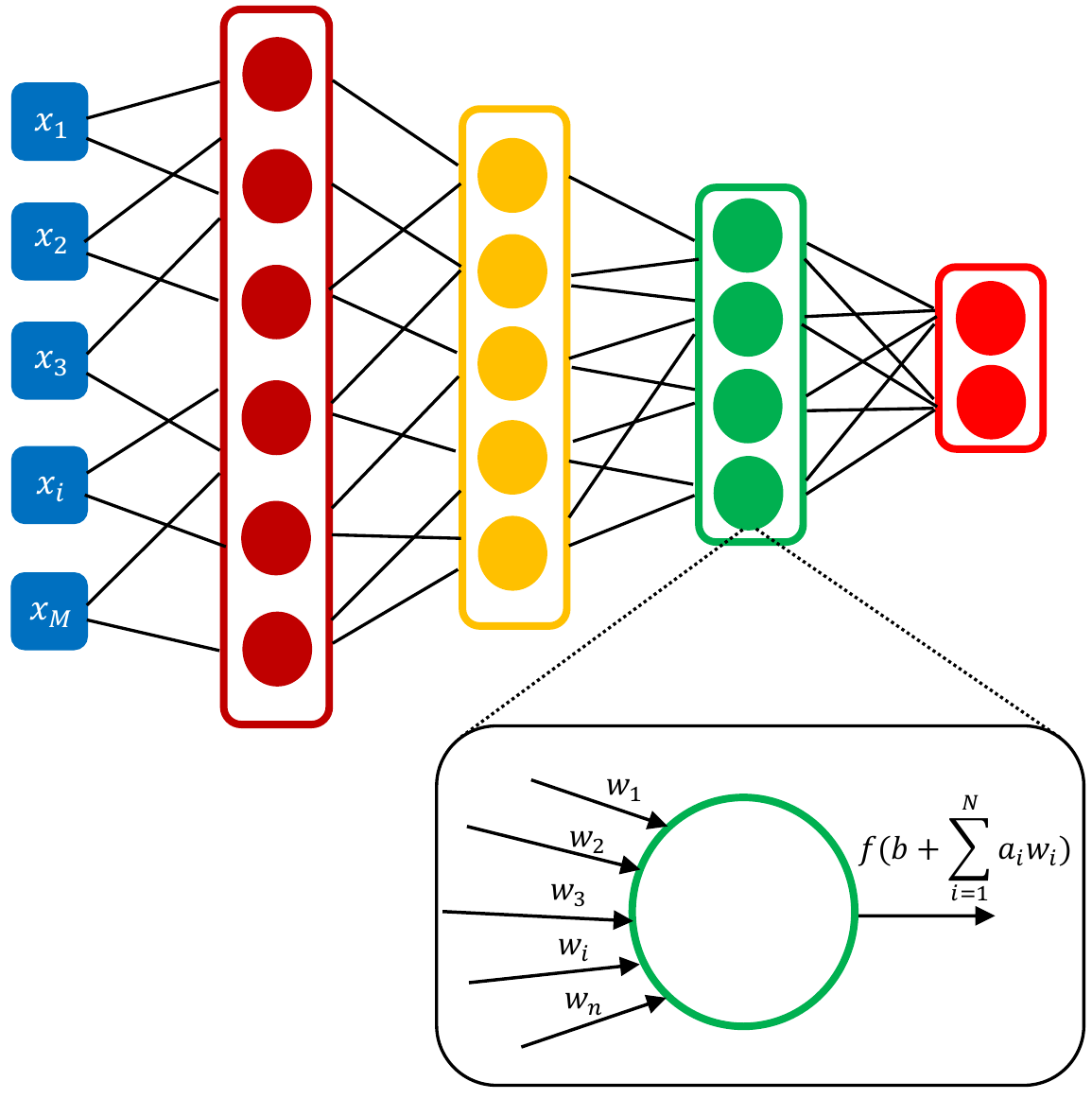}
	\caption{Multi-layer Perceptron}
	\label{mlp}
\end{figure}

In stage 1, our objective is to predict the availability of services at location $l_i$ and time $t_i$ based on their previous spatio-temporal logs i.e. the historical time-geolocation of their availability. Our prediction model is based on several key features inferred from raw data including spatio-temporal attributes. This stage is composed of two levels: Offline Processing and Online Processing. At the first level, a deep learning model is trained on the data features. Once trained, the model can be queried online in real-time to predict the available crowdsourced services at location $l_i$ and time $t_i$.   \newline
We explicitly define the following set of features: 
\begin{itemize}
	\item Latitude $\phi$ in WGS84 decimal degree.
	\item Longitude $\tau$ in WGS84 decimal degree.
	\item Time of day $t$ with high granularity: hour, minute, second. 
	\item Day of the week $d$. 
	\item Is Weekday $wd$: a variable that indicates the type of the day: 
		\begin{equation}
			\begin{cases}
			1, & \text{if $d$ is a weekday (Monday, \ldots, Friday)}\\
			0, & \text{if $d$ is a weekend (Saturday, Sunday)}
			\end{cases}
		\end{equation}
	\item Is Holiday $hd$: a variable indicating if the day is a holiday: 
		\begin{equation}
			\begin{cases}
		1, & \text{if $d$ is an official holiday}\\
		0, & \text{otherwise}
			\end{cases}
		\end{equation}
	
\end{itemize}
The core component of the crowdsourced service prediction is the deep neural network (DNN). This network has  multiple layers of neurons representing the network depth. Unlike the shallow models, deep networks are generally characterized by higher depth. Following the recent research breakthroughs in machine learning, training such network has become significantly easier. DNNs can model high complex non-linear relationships. Each layer extracts higher level features from the previous layers. We rely in our framework on the multi-layer perceptron model (MLP) depicted in Fig. \ref{mlp}.
MLP is the quintessential deep learning model. It is of extreme importance and represents the basis for many commercial applications \cite{ref14}. MLP seeks to approximate a function $f^*$. $f^*$ maps a data point $x$ to its class label in case of a classifier. The feedforward network applies the mapping $y=f(x,\theta)$ and determines the optimal set of parameters $\theta$ that gives the best data approximation. Therefore, a network is a composition of many different functions. An MLP has at least three layers of nodes which means $f(x) = f_1(f_2(f_3(x)))$. In this case, a function represents a layer i.e. $f_1$ is the first layer, etc.  Node, or perceptron, represents the baseline of DNN. At the perceptron level, data are non-linearly transformed using an activation function inspired from neuroscience. However, modern activation functions are instead derived from mathematical disciplines.  In this first stage, we opt for a Leaky ReLU \cite{ref16} activation given by: 
\begin{equation}
\text{Leaky\_ReLU(x)}=\begin{cases}
1, & \text{if $x\geq0$}.\\
ax, & \text{otherwise}.
\end{cases}
\label{lrelu}
\end{equation}
Where $a$ is a fixed positive parameter less than 1 usually recommend to be very small  (e.g. 0.01). Leaky ReLU is an attempt to fix the dying ReLU (where $a = 0$) problem. Indeed, ReLU can be fragile and dies during training. In other words, a large gradient can result in weights update in such a way that the neuron will no longer activate on future data sample. \newline
The learning process consists of updating the set of parameters $\theta$ based on an amount of error modeled by an objective function or loss. This process is called backpropagation \cite{ref16,ref_ahmed}. The mean square error is a classic loss choice, generally used for regression problem: 
\begin{equation}
J(x,\theta)=||f^*(x) -f(x,\theta)||^2
\end{equation}
A typical choice of loss for classification problem is the binary cross-entropy, defined for $N$ training samples as: 
\begin{equation}
	J(x,\theta)=-\frac{1}{N}\sum_iy_ilog(f^*(x_i))+(1-y_i)log(1-f^*(x_i))
	\label{crossent}
\end{equation}
 The optimization of $J((x,\theta))$ is conducted using an optimizer such as the stochastic gradient descent where the update of  parameters is carried out using the following routine: 
\begin{equation}
	\theta = \theta -\psi\nabla_\theta J(\theta;x_i)
	\label{backprop}	
\end{equation} 
Where $\nabla$ is the gradient operator and $\psi$ is the learning rate. Furthermore, we apply a batch normalization procedure between layers to ensure faster learning and higher prediction accuracy. Algorithm \ref{algo1} gives details of the batch normalization. First, the mean and variance of the batch are calculated. Then the batch is normalized by centering (subtracting the mean) and scaling to unit variance. An $\epsilon > 0$ is used to avoid dividing by zero. Finally the batch is scaled and shifted.
\begin{algorithm}
	\caption{Batch Normalization (BN) }\label{algo1}
	\hspace*{\algorithmicindent} \textbf{Input}: Batch $B =\{X_i, …, X_m\}$  \\
	\hspace*{\algorithmicindent} \hspace*{\algorithmicindent} \hspace*{\algorithmicindent} Parameters: $\alpha$, $\beta$\\
	\hspace*{\algorithmicindent} \textbf{Output}: Normalized batch $\{Y_i = BN_{\alpha,\beta}(X_i)\}$   \\

	\begin{algorithmic}

		\State \textit{\#Batch Mean}
		\State
		\State $\boldsymbol{\mu = \frac{1}{N}\sum_i X_i}$
		\State
		\State \textit{\#Batch Variance}
		\State
		\State $\boldsymbol{\sigma^2 = \frac{1}{N}\sum_i (X_i-\mu)^2}$
		\State
		\State \textit{\# Normalize}
		\State
		\State $\boldsymbol{\hat{X_i}=\frac{X_i-\mu}{\sqrt{\sigma^2 +\epsilon}}}$
		\State
		\State \textit{\#Scale and Shift}
		\State
		\State $\boldsymbol{Y_i=\alpha\hat{X_i}+\beta}$

	\end{algorithmic}

\end{algorithm}\newline
Algorithm \ref{algo2} outlines the training process of the crowdsourced service availability prediction. First, we determine the cluster of each pair of crowdsourced service $X_i$ and its associated spatio-temporal features $E_i$. Next, a batch of crowdsourced services, spatio-temporal features, cluster ids and service ids is fed into the network model layer by layer using Eq. \ref{lrelu}. The output of each layer is also normalized using \ref{algo1}. At the last layer, the objective function is evaluated and minimized in order to update the network parameters. This routine is repeated till a stopping criterion is satisfied.\newline
\begin{algorithm}
	\caption{Crowdsourced service availability prediction }\label{algo2}
	\hspace*{\algorithmicindent} \textbf{Input}: Historical Crowdsourced services spatiotemporal data $\{X_1,X_2,\cdots X_N\}$  \\
	\hspace*{\algorithmicindent} \hspace*{\algorithmicindent} \hspace*{\algorithmicindent}Other features $E$: month, day, hour, minute, second, is\_weekend, is\_weekday, \hspace*{\algorithmicindent} \hspace*{\algorithmicindent} \hspace*{\algorithmicindent} is\_holiday\\
	\hspace*{\algorithmicindent} \textbf{Output}: learned DNN model to predict the availability of services at location $l_i$ and time $t_i$  \\
	
	\begin{algorithmic}
		
		\State \textit{\# Construct training instance}
		\State \textbf{For every pair of {Xi, Ei}} 
		\State  \indent Find the cluster $C_i$ to which $X_i$ belongs 
		\State \indent Create instance $\boldsymbol{\{X_i, E_i, C_i, Sid\}}$
		\State \textit{\#Train the model}
		\State \textbf{Repeat}
		\State \indent Select batch of instances $\boldsymbol{\{X_i, E_i, C_i, Sid\}}$
		\State \indent \textbf{For every layer}
		\State \indent \indent Feed the batch using \textbf{(\ref{lrelu})} 
		\State \indent \indent Apply Batch Normalization using \textbf{Algorithm \ref{algo1}} 
		\State \indent Find $\theta$ that minimizes \textbf{(\ref{crossent})} using \textbf{(\ref{backprop})}
		\State \textbf{Until} stopping criterion is met
		
	\end{algorithmic}
	
\end{algorithm}
After training and obtaining the optimal parameters, the DNN can be queried by providing the geolocation and time associated to the query. The DNN predicts the corresponding service id(s). 

\subsection{Stage 2: Crowdsourced service duration prediction }
In this stage, we aim at finding the temporal attributes of the available services predicted at stage 1. Indeed, it is important to determine these attributes to enable any potential application such as task assignment and coverage service composition. Therefore, it is essential to develop an accurate prediction model. We propose to formulate the task as time series forecasting to estimate the duration of service availability. Our approach is to construct the time series of every service provider associated to every cluster and sampled with a particular time granularity. The forecast is based on predicting whether the service is present at this particular location or not. The forecast can be conducted for multi-step ahead i.e. the next $\gamma$ times. In the upcoming sub-sections, we describe the process of time series generation and how to transform these times series into two types of Gramian Angular Field images. We also present our DNN architecture for time series forecasting.
\subsubsection{Time series generation}
Initially, an offline pre-processing of the historical data is conducted. Specifically, for every crowd service provider, we construct the time series of appearance at a particular cluster by assigning the historical geolocations to the set of clusters {$C_{1}$, $C_{2}$, …, $C_{N}$}. Let $TS$ be a $T$ timestamp vector sequence $TS= (ts_1, ts_2, \cdots, ts_T)$. $ts_i$ is given by: 
\begin{equation}
ts_i=\begin{cases}
1, & \text{if $S_{id} \in C_j$ at time $i$ }\\
0, & \text{otherwise}.
\end{cases}
\label{tseries}
\end{equation}
Our forecasting approach is based on rolling and encoding time series as images. This allows applying recent deep learning techniques of high effectiveness  particularly in computer vision applications. These techniques enable learning different image patters and structure. To image the time series, first, we run a rolling window operation on the data. For each  time series of $n$ values, a window $w$ of length $k \leq n$ and an overlapping ratio $r$, we take the first block of length $k$ and then roll the window by $r$. Therefore, the second block of $k$ data observation starts from the $r_{th}$ observation.  By doing so, we are looking at the changing property of the time series over time instead of one single observation.  \newline
The forecasting problem is a multi-step ahead prediction i.e. we forecast the service presence at a given cluster for the next $\gamma$ times (3 minutes for example). Suppose that the time series is sampled per minute and $\gamma=3$, the forecasting process consists of predicting the label $L = [l_1, l_2, l_3]$ where $l_i \in {0, 1}$. Therefore, the time series forecasting is formulated as a multi-class classification task where the the number of classes is $2^\gamma$. 
\subsubsection{Imaging time series}
\begin{figure}[!h]
	\centering
	\includegraphics[scale=0.3]{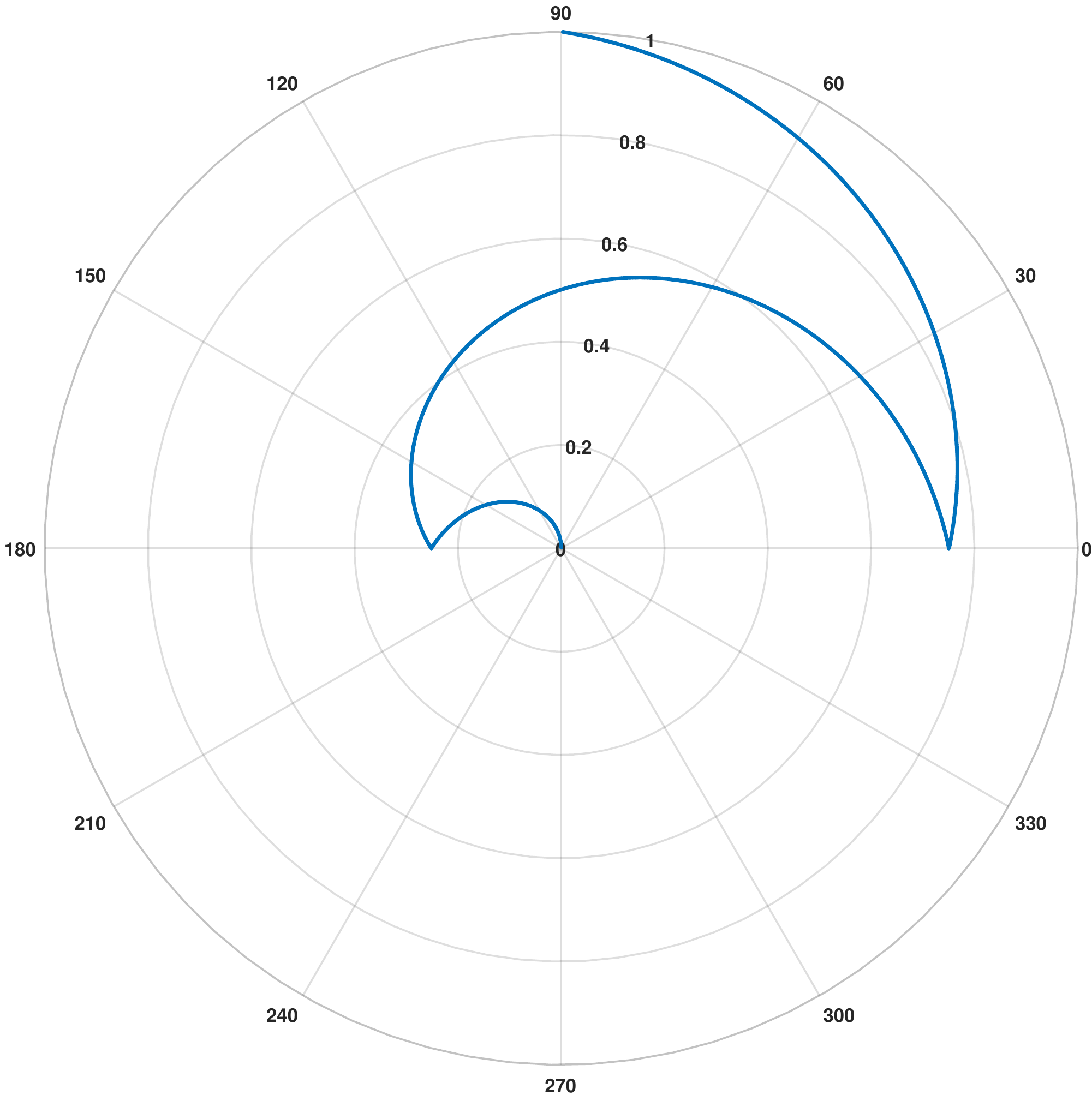}
	\caption{Polar plot of a sinusoidal time series}
	\label{polar_sin}
\end{figure}

\begin{figure}[!h]
	\centering
	\includegraphics[scale=.6]{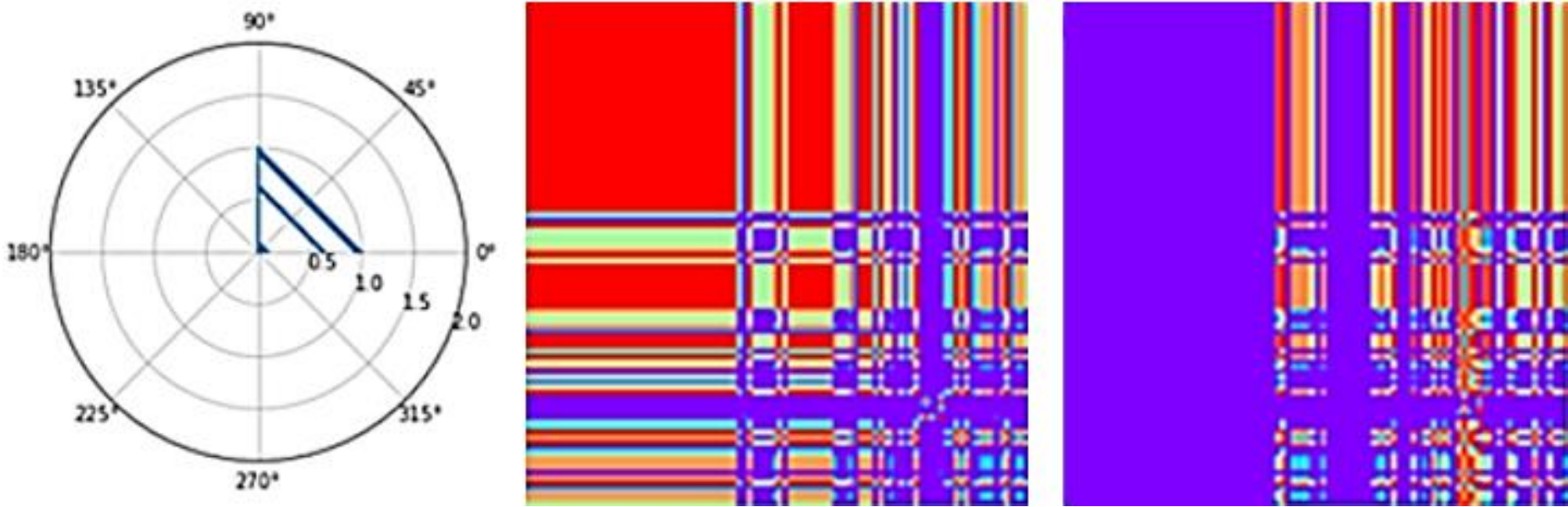}
	\caption{Binary time series: Polar plot(left), GASF (middle) and GADF (right)}
	\label{binary_ts}
\end{figure}

To image the time series blocks, we propose to use the Gramian Angular Field \cite{ref17}. Specifically, the time series is presented in polar coordinates system instead of the classic Cartesian system. Each element of the Gramian matrix is the cosine of the summation of two angles. A time series $TS= (ts_1, ts_2, …, ts_n)$, $ts_i$ can be transformed to polar coordinates. This operation is conducted by encoding $ts_i$ as angular cosine value and its corresponding timestamp $i$ as the radius: 
\begin{equation}
	\begin{tabular}{c}
	\(\Psi_i=arccos(ts_i); \quad ts_i \in TS \) \\
	\( \rho_i = \frac{i}{C}; \quad i \in \mathbb{N} \)
	\end{tabular}
\end{equation}
Where the constant $C$  allows controlling the polar coordinates system span.
For example, we illustrate in Fig. \ref{polar_sin} the polar plot of a sinusoidal time series. \newline
After rescaling and transforming the time series to the polar coordinates system, the angular property can be exploited by trigonometric means. The idea is to exploit the correlation in time. The Gramian Angular Summation Field (GASF) is a matrix of the form:
\begin{equation}
	\begin{tabular}{c}
				\(GASF_{i,j}=cos(\Psi_i+\Psi_j) \)\\
				\( =tr\big(TS\big).TS - tr\big( \sqrt{\big( I-TS^2 \big)}. \big(  I-TS^2\big)  \big)  \)
	\end{tabular}
\end{equation} 
$GASF_{i,j;|i-j|=k}$ represents the correlation by superposition of directions given time $k$.
The Gramian Angular Difference Field (GADF) is a matrix defined as:
\begin{equation}
\begin{tabular}{c}
\(GADF_{i,j}=sin(\Psi_i+\Psi_j) \)\\
\( =tr\big( \sqrt{\big( I-TS^2 \big)}.TS - tr\big(TS\big). \big( \sqrt{\big( I-TS^2 \big)} \)
\end{tabular}
\end{equation} 
$GADF_{i,j;|i-j|=k}$ represents the correlation by difference of directions given time $k$. 
$I$ is the unit vector  and $tr(X)$  is the transpose of $X$. The generation of GASF and GADF results in matrices of the size $T\times T$. If the size is too large for manipulation, one can easily apply the Piecewise Aggregation Approximation \cite{ref18} to reduce the time series size without losing its trend. For a binary time series, we illustrate in Fig. \ref{binary_ts}, its polar plot, GADF and GASF. 
\subsubsection{Deep Neural Network for multi-step ahead time series forecasting }
\begin{figure}[!h]
	\centering
	\includegraphics[scale=.6]{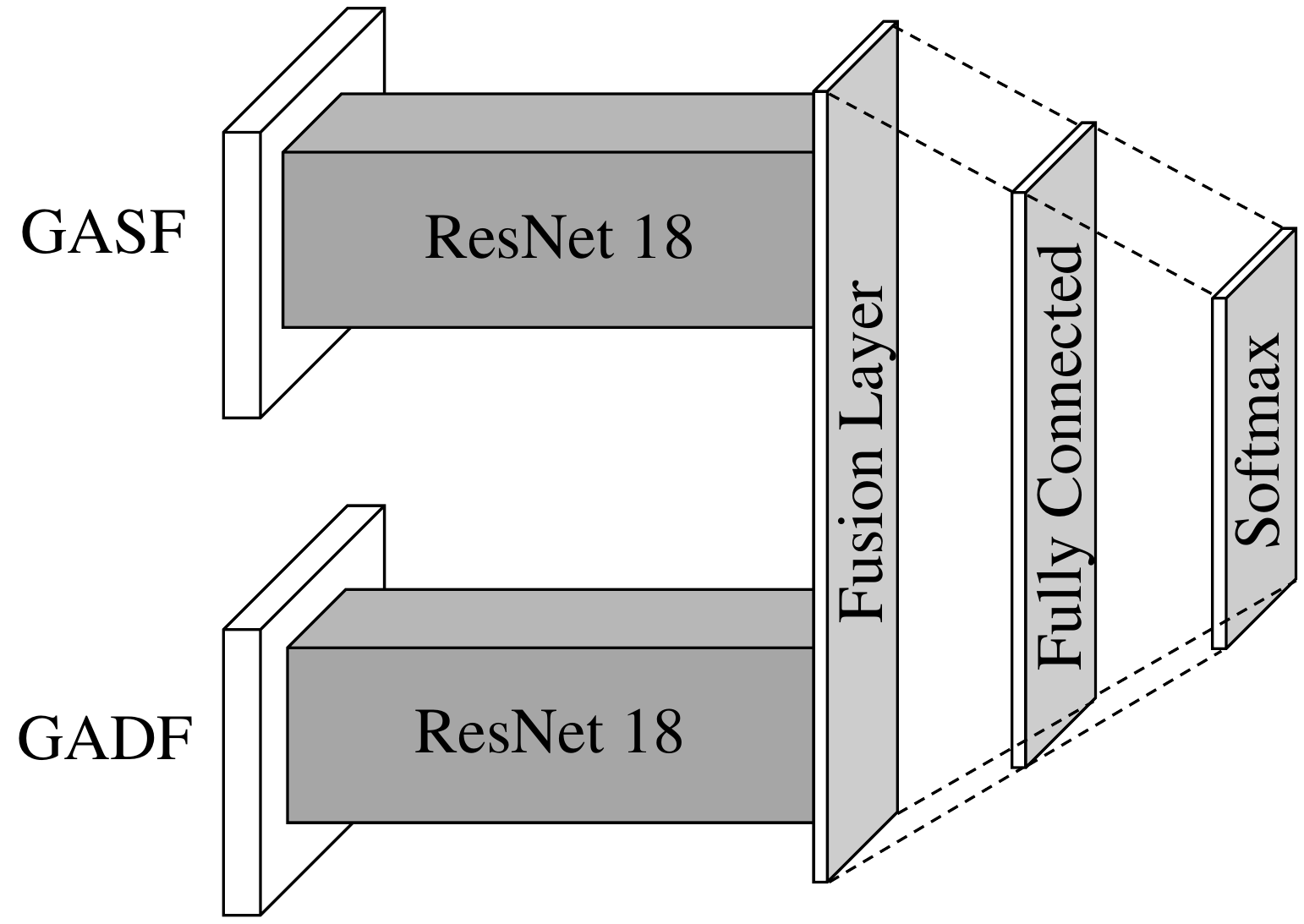}
	\caption{Time series forecasting model}
	\label{2nd_model}
\end{figure}

\begin{figure}[!h]
	\centering
	\includegraphics[scale=.8]{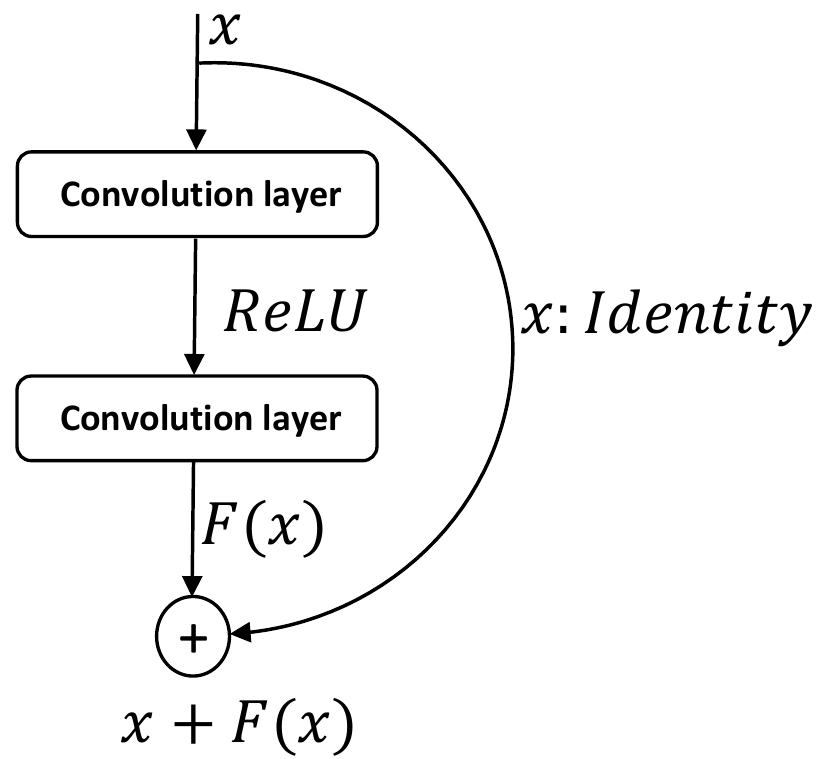}
	\caption{Resnet building block}
	\label{resnet}
\end{figure}

We propose to exploit both the GASF and GADF to accurately forecast time series. The general 2$^{nd}$ stage DNN architecture is illustrated in Fig. \ref{2nd_model}.
It consists of two ResNet-18 \cite{resnet} pathways for both GAF images. At each pathway, the image is fed into ResNet-18, a particular type of deep network. The outputs of each pathway are then concatenated at the fusion layer and then fed into a fully connected layer. The Softmax layer enables distinguishing between each input based on its class as classes are mutually exclusive. This is done by applying the Softmax function on the previous layer. The Softmax represents a reasonable alternative to the max function as it is differentiable and outputs maximum probability for the maximum value of the previous layer and 0 for the rest. ResNet-18 is an 18-layer DNN of residual Convolution Neural Network (CNN) \cite{cnn1,cnn2}. CNN is a particular type of neural network that performs convolution operation of a $n\times m$ filter $w$ on tensor input $a$. For the $k$ $l^{th}$ hidden neuron, the convolution operation is:
\begin{equation}
	f(  \sum_i^{n-1} \sum_j^{m-1} w_{i,j}a_{k+i,l+j}  +b )
	\label{cnn}
\end{equation} 
Residual network offers a simple yet efficient way to train very deep neural network without suffering the notorious problem of vanishing/exploding gradient \cite{exploide}. The residual block maps the input $x$ to $F(x)$  through two layers of neural networks (CNN for instance) with ReLU activation. At the output, the input $x$ is summed with $F(x)$. This is done through an identity skip connection that carries out the input $x$ to the output. The residual block is illustrated in Fig. \ref{resnet}. 
The purpose of going deeper in neural network is to learn something new from previous layer. By providing the input without transformation, we are driving the new layer to learn something new and different from what has been already encoded in previous layers. With such building block, it is safe to go deeper since in worst case scenario, the unnecessary layers will run an identity mapping and will not degrade performance. \newline
ResNet-18 network is detailed in Tab. \ref{tab_resnet}. The input image is fed into a $7 \times 7$ CNN of stride 2 which is connected to a max pooling layer (downsampling by max operation) to reduce image size. Next, it is propagated in 8 blocks of residual network of 64, 128, 256 and 512 CNN filters connected to an average pooling layer (downsampling by average operation), a fully connected layer and a Softmax to output the image label. In our proposed architecture, the ResNet-18 Softmax and Fully Connected layers are dropped and a fusion layer is added where the output of the average pooling layer of both pathways are concatenated. This layer is connected to a 2$^\gamma$-d Fully connected layer linked to a Softmax. 
\begin{table}[h!]
	\begin{center}
		\caption{ResNet-18}
		\label{tab_resnet}
		\begin{tabular}{|*2{>{\renewcommand{\arraystretch}{0.5}}c|}} 
			\hline
			$7 \times 7$ 64 stride 2 \\
			\hline
			$3 \times 3$ max pool stride 2 \\			
			\hline
			$ \left[ \begin{array}{c} 3 \times 3, 64    \\ 3 \times 3, 64  \end{array}\right]$  $\times$ 2\\
						\hline
			$ \left[ \begin{array}{c} 3 \times 3, 128    \\ 3 \times 3, 128  \end{array}\right]$  $\times$ 2\\
						\hline
			$ \left[ \begin{array}{c} 3 \times 3, 256    \\ 3 \times 3, 256  \end{array}\right]$  $\times$ 2\\
						\hline
			$ \left[ \begin{array}{c} 3 \times 3, 512    \\ 3 \times 3, 512  \end{array}\right]$  $\times$ 2\\
			\hline
			Average pooling, 
			2$^\gamma$-d Fully\\
			 Connected, Softmax\\
			\hline			
		\end{tabular}
	\end{center}
\end{table}
In addition, to maximize the prediction performance, we use a learning scheduler. Indeed, it is well-known that the convergence of stochastic gradient descent is carried out by reducing the step size. However, a scheduling approach where the learning rate is changed over time, can lead to significantly faster convergence to better minimum. Initially, a large step leads to fast minimization of the objective function. Later, smaller steps are necessary in order to reach finer minimum of the loss. In our training algorithms, we use the following scheduler: 
\begin{equation}
	\alpha=\alpha . \delta^{floor(epoch/drop)}\label{drop}
\end{equation}
Where $\delta$ is the drop rate, $drop$ is the epoch drop and $floor(x)$ is the function that returns the integer less  than or equal to $x$. The second term controls the amount of the learning rate decrease from the previous epoch.\newline
The pseudocode of training the DNN of the time series forecasting is detailed in Algorithm \ref{algorithm3}.
Once the model is trained, it can be queried to predict in the next $gamma$ times the presence of the corresponding crowdsourced service in this particular geolocation and particular time. 
\begin{algorithm}
	\caption{Time series forecasting }\label{algorithm3}
	\hspace*{\algorithmicindent} \textbf{Input}: $X_{GASF}$, $X_{GADF}$  \\
	\hspace*{\algorithmicindent} \textbf{Output}: learned DNN forecasting model   \\
	
	\begin{algorithmic}
		\State \textbf{Repeat}
		\State \textit{\# For both GASF and GADF pathways}
		\State \indent Feed the data using \textbf{(\ref{cnn})} in $\boldsymbol{7 \times 7, 64}$, stride \textbf{2} CNN
		\State \indent Apply max pooling with $3 \times 3$ filter
		\State \indent Feed the data in two residual blocks of $\boldsymbol{3 \times 3, 64}$ CNN
		\State \indent Feed the data in two residual blocks of $\boldsymbol{3 \times 3, 128}$ CNN
		\State \indent Feed the data in two residual blocks of $\boldsymbol{3\times 3, 256}$ CNN
		\State \indent Feed the data in two residual blocks of $\boldsymbol{3\times 3, 512}$ CNN
		\State \indent Apply Average pooling
		\State \indent \textit{\# Fusion layer}
		\State \indent Concatenate outputs of Average pooling layers
		\State \indent Feed data using \textbf{(\ref{lrelu})} in $\boldsymbol{2^\gamma}-d$ Fully Connected
		\State \indent Apply Softmax
		\State \indent Find $\theta$ that minimizes \textbf{(\ref{crossent})} using \textbf{(\ref{backprop})}
		\State \indent Decreas $\boldsymbol{\alpha}$ using \textbf{(\ref{drop})}
		\State \textbf{Until} Stopping criterion is met
		
	\end{algorithmic}
	
\end{algorithm}

\section{Experimental results}
To assess the performance of the prediction framework, we run  multiple experiments on different datasets. Due to the lack of crowdsourced data, we use alternative datasets with spatio-temporal records: 
\begin{itemize}
	\item ECML/PKDD-15 dataset \cite{taxi} provides one year records (from 01-07-2013 to 30-06-2014) of trajectories for 442 taxis operating in Porto, Portugal. For every taxi, the id, geolocations sampled every 15 seconds, timestamp and day type are extracted. We also eliminate taxi ids which have few occurrences in the data. This results in 428 taxi ids.
	\item Foursquare dataset \cite{fourquare} provides check-in data   for around 10 months in  New York collected from April 2012 to February 2013. These check-ins correspond to 1083 users on 400 categories of venues. For every check-in, several features are provided. We keep for our experiments the user id, latitude, longitude and timestamp. We also eliminate the user ids with few check-in records.
	\item Uber dataset \footnote{https://www.kaggle.com/theoddwaffle/uber-data-analysis/data} contains information about over 4.5 million Uber pickups from April to September 2014, and 14.3 million more Uber pickups from January to June 2015  in New York City. The dataset includes also Trip-level data on 10 other for-hire vehicle (FHV) companies in addition to aggregated data for 329 FHV companies. We conduct our experiments on Uber data records of 5 service providers tracked in April 2014.
\end{itemize}
 The gap statistics is applied to derive the optimal number of clusters. Our findings showed that C$_{ECML/PKDD-15}$ = 8, C$_{Foursquare}$=12 and C$_{Uber}$ = 3. 
 
 \subsection{Crowdsourced service availability prediction}
\begin{figure}[!h]
 	\centering
 	\includegraphics[scale=.5]{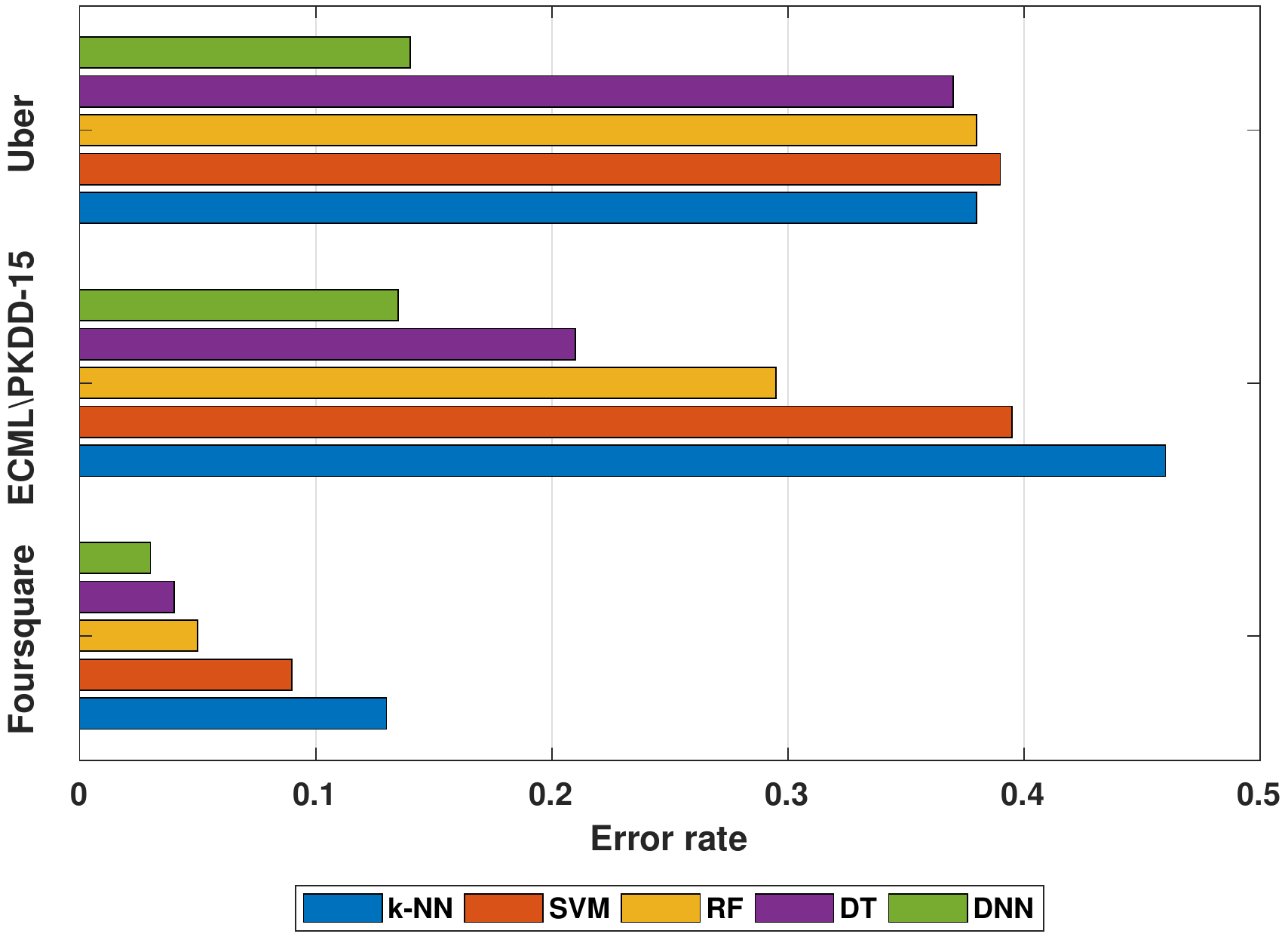}
 	\caption{Service availability prediction: error rate}
 	\label{serv_pred_error}
 \end{figure}

\begin{figure}[!h]
	\centering
	\includegraphics[scale=.5]{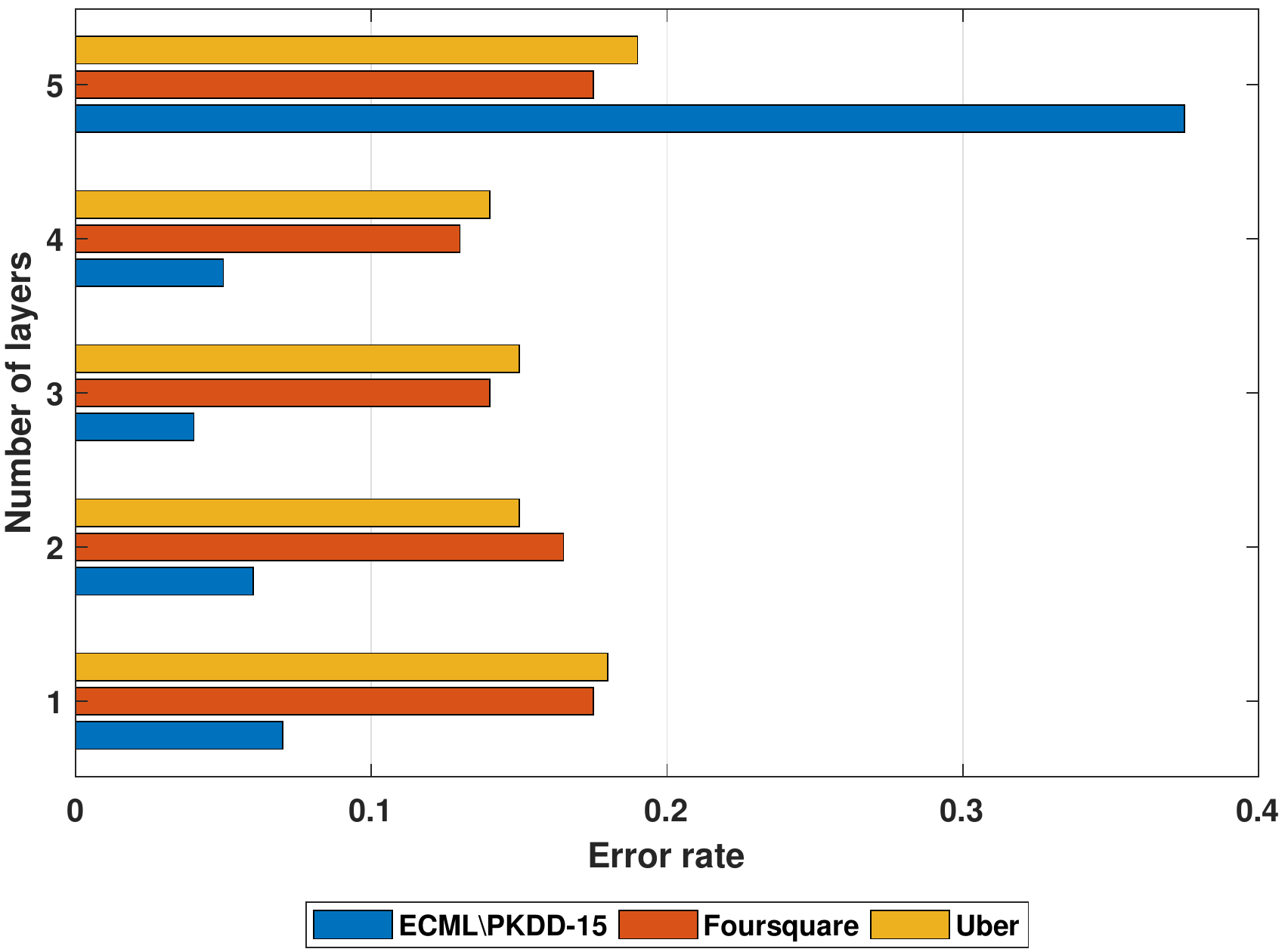}
	\caption{Service availability prediction: error rate vs number of hidden layers}
	\label{layers_error}
\end{figure}

 We evaluate the deep learning-based crowdsourced service prediction approach against well-known classic algorithms: Random Forest (RF), Decision Tree (DT), k-Nearest Neighbors (kNN) and Support Vector Machine (SVM). We experiment with different parameters for every algorithm and we report the best result.\newline
 We split all datasets into 80/20 training/testing with 10\% of the data reserved for validation. We choose batch size = 256 for ECML/PKDD-15, batch size = 128 for Foursquare and Uber. We train the DNN until the decrease in error is less than $10^{-4}$. We illustrate in Tables \ref{tab_pred1}, \ref{tab_pred2} and \ref{tab_pred3} the deep learning architecture for each dataset. The ECML/PKDD-15 DNN has 4 layers where the first two layers have 512 neurons activated using Leaky-ReLU with $a = 0.01$ while the last two layers have 448 neurons activated using Leaky-ReLU with $a = 0.02$. For Foursquare dataset, we use a 5-layer DNN. The first three layers have 256 neurons while the last two layers have 128 neurons. All layers are activated using Leaky-ReLU with $a = 0.01$. The Uber prediction DNN has 3 Leaky-ReLU layers with $a=0.01$. The first and second layers have 16 neurons while the third layer has 8 neurons. For all DNNs, each layer is connected to a Batch Normalization layer. We empirically set the value of parameter $a$ i.e. the ones leading to the best performance. All DNNs were trained on a GTX-670MX of 3GB memory and required 15 minutes for ECML/PKDD-15, 12 minutes for Foursquare dataset and 10 minutes for Uber dataset. 
 \begin{table}[h!]
 	\begin{center}
 		\caption{ECML/PKDD-15 prediction DNN}
 		\label{tab_pred1}
 		\begin{tabular}{|*2{>{\renewcommand{\arraystretch}{0.5}}c|}} 
 			\hline
 			$ \left[ \begin{array}{c} 512 \, \text{Leaky-ReLU}, a=0.01    \\ \text{Batch} \, \text{Normalization()}   \end{array}\right]$  $\times$ 2\\
 			\hline
 			$ \left[ \begin{array}{c} 448 \, \text{Leaky-ReLU}, a=0.02    \\ Batch \, Normalization ()  \end{array}\right]$  $\times$ 2\\
 			\hline
 			428 Softmax\\
 			\hline			
 		\end{tabular}
 	\end{center}
 \end{table}
  \begin{table}[h!]
 	\begin{center}
 		\caption{Foursquare prediction DNN}
 		\label{tab_pred2}
 		\begin{tabular}{|*2{>{\renewcommand{\arraystretch}{0.5}}c|}} 
 			\hline
 			$ \left[ \begin{array}{c} 256 \, \text{Leaky-ReLU}, a=0.01    \\ \text{Batch} \, \text{Normalization()}  \end{array}\right]$  $\times$ 3\\
 			\hline
 			$ \left[ \begin{array}{c} 128 \, \text{Leaky-ReLU}, a=0.01    \\ \text{Batch} \, \text{Normalization()}  \end{array}\right]$  $\times$ 2\\
 			\hline
 			50 Softmax\\
 			\hline			
 		\end{tabular}
 	\end{center}
 \end{table}
   \begin{table}[h!]
 	\begin{center}
 		\caption{Uber prediction DNN}
 		\label{tab_pred3}
 		\begin{tabular}{|*2{>{\renewcommand{\arraystretch}{0.5}}c|}} 
 			\hline
 			$ \left[ \begin{array}{c} 16 \, \text{Leaky-ReLU}, a=0.01    \\ \text{Batch} \, \text{Normalization()} \end{array}\right]$  $\times$ 2\\
 			\hline
 			$ \left[ \begin{array}{c} 8 \, \text{Leaky-ReLU}, a=0.01    \\ \text{Batch} \, \text{Normalization()}  \end{array}\right]$\\
 			\hline
 			5 Softmax\\
 			\hline			
 		\end{tabular}
 	\end{center}
 \end{table}
We report in Fig.\ref{serv_pred_error} the classification error rate on the test sets for each dataset. These results show that DNN based approach achieved the lowest classification error rate on all datasets. For example, on Foursquare data, RF and DF achieved an average error rate of 0.21 and 0.29 respectively. SVM and k-NN failed to achieve a good service prediction as the error rate was 0.39 and 0.46. However, the DNN showed a good prediction accuracy with an error rate of 0.13.  These findings confirm that by opting for a deep learning approach, we can achieve the best crowdsourced service prediction performance. By going deeper, we can learn multiple levels of abstractions from the data. For this reason, we further analyze how the number of layers affects the DNN prediction performance. Results are detailed in Fig. \ref{layers_error}. We notice that, initially, the error rate decreases as the DNN goes deeper but starts to increase. Indeed, the model starts overfitting i.e. performing  well during the training phase but less efficient in the testing phase which would require regularizing the network by including dropout layers \cite{dropout} for example. In addition, by adding more layers, we increase the model complexity which requires more training data to reach good generalization. 

\subsection{Crowdsourced service duration prediction}
\begin{figure}[!h]
	\centering
	\includegraphics[scale=.5]{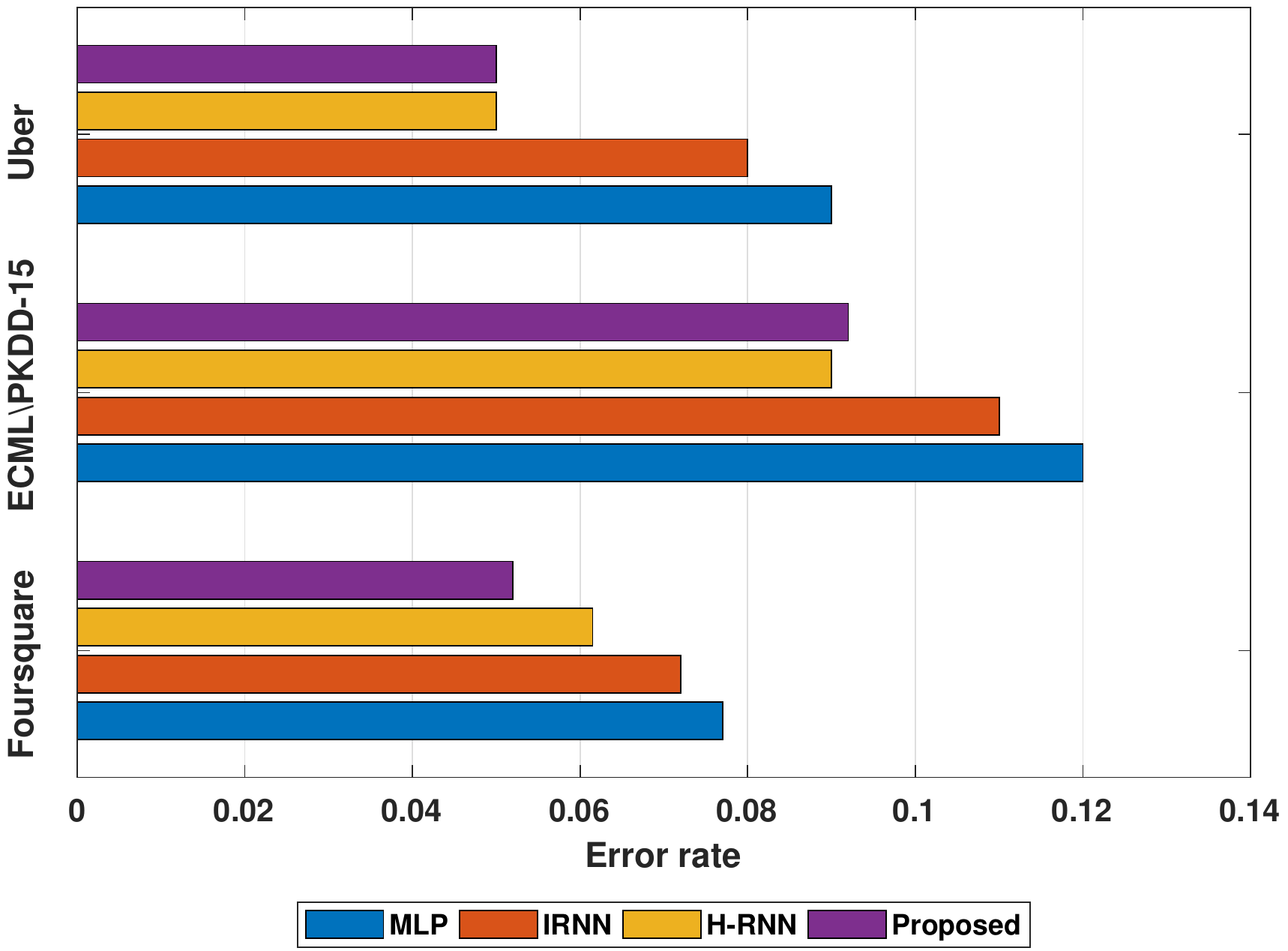}
	\caption{Forecasting error rate for one minute ahead}
	\label{one_min}
\end{figure}

\begin{figure}[!h]
	\centering
	\includegraphics[scale=.5]{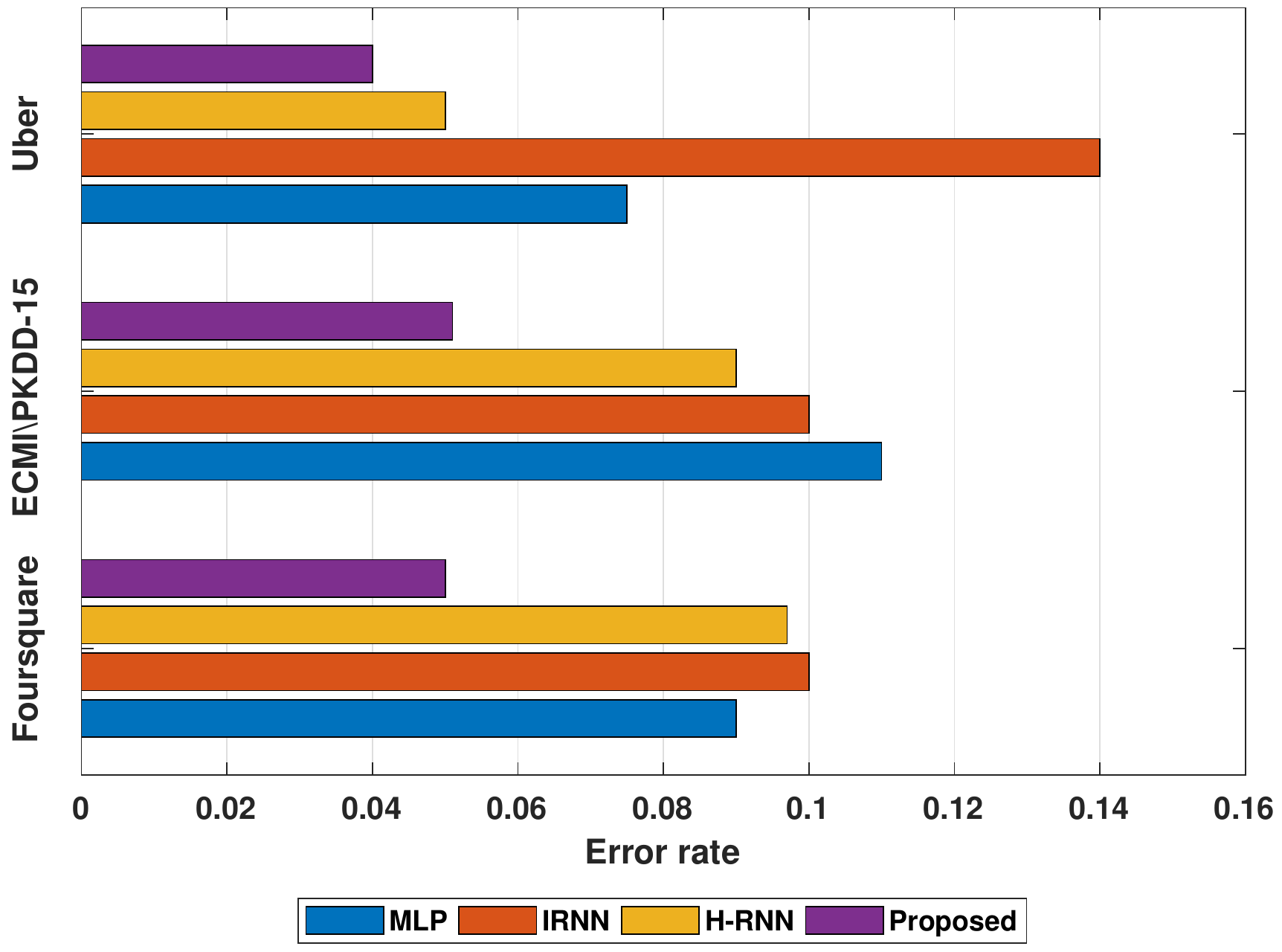}
	\caption{Forecasting error rate for two minutes ahead}
	\label{two_min}
\end{figure}

\begin{figure}[!h]
	\centering
	\includegraphics[scale=.5]{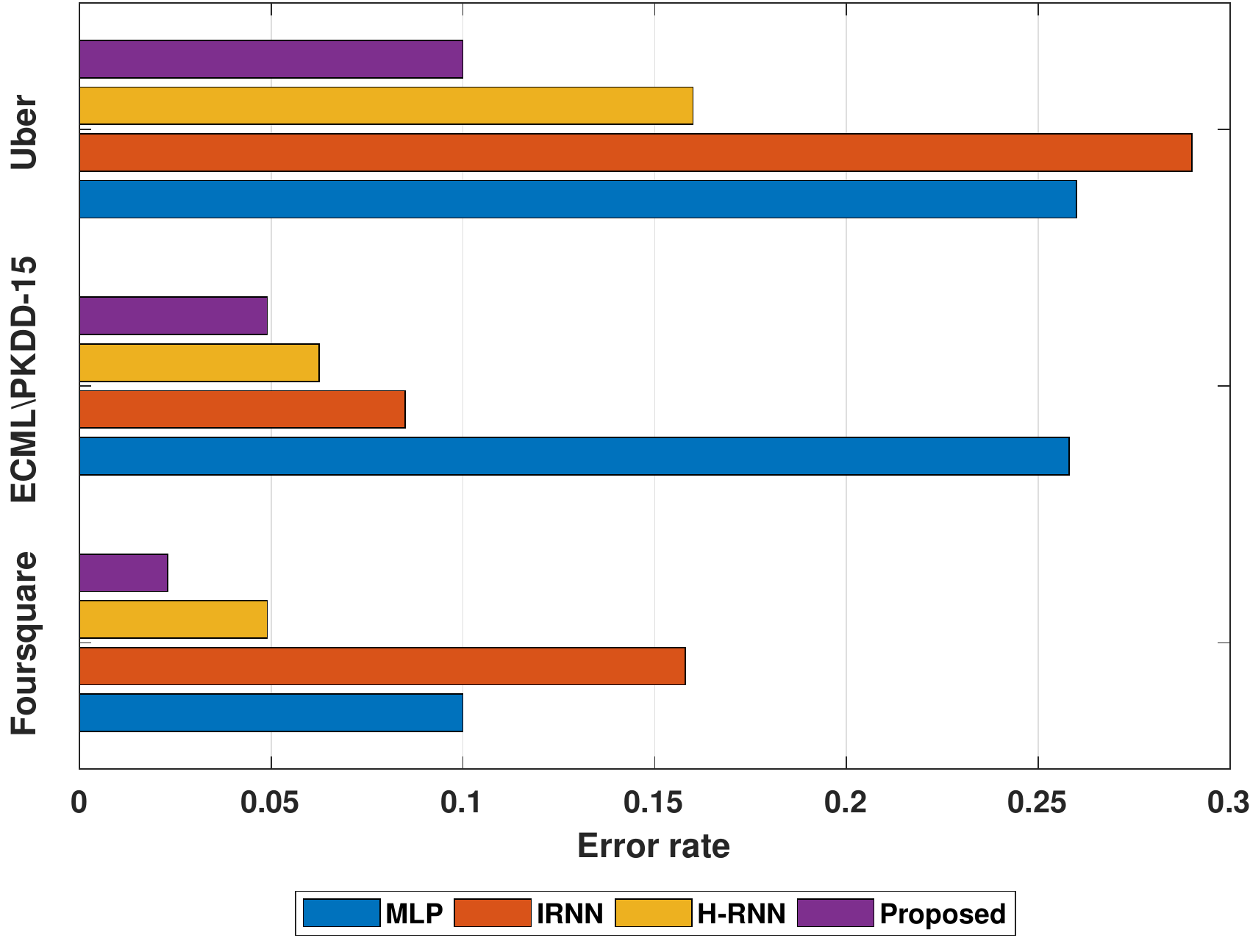}
	\caption{Forecasting error rate for three minutes ahead}
	\label{three_min}
\end{figure}

Once the service availability is predicted at the first stage and based on the binary time series of its presence at the corresponding location, the second stage DNN model depicted in Fig. \ref{2nd_model} predicts at multiple step ahead the availability duration of the service. In our experiment, we set $\gamma=$1, 2 and 3 that is we conduct experiments to forecast the service presence for the next one minute, next two minutes and next three minutes. Therefore, the set of labels are: $\{0, 1\}_{\gamma=1}$, $\{00, 10, 01, 11\}_{\gamma=2}$ and $\{000, 100, 010, 110, 001, 101, 011, 111\}_{\gamma=3}$. We also add a small perturbation $\epsilon = 10-3$ to the zeros time series to ensure proper construction of the GASF and GADF. For both datasets, we set up the learning scheduler parameters as: $\alpha=0.1$, $ \tau= 0.5$ and drop = 10. \newline
The data turn out to be unbalanced for all the step ahead setting. Indeed, the most dominant forecasting label is $\{0\}$ since it is less likely one service will be present in the same location every day. This unbalanced property would lead to poor prediction performance for all DNN based approaches. Therefore, we conduct a data augmentation procedure where we generate artificial data by manipulating the GASF and GADF images. Specifically, we rotate the image with an angle = 40 degree and apply a shearing transformation on the image. By doing so, we are able to balance the training data in order to ensure good generalization on the test set.\newline
We compare the proposed ResNet-based DNN for time series forecasting with the hierarchical Recurrent Neural Network \cite{hrnn} applied instead of ResNet on the same architecture. We also provide comparison with two forecasting DNNs applied on the raw data (without GASF and GADF transform) where we use in the first one IRNN \cite{irnn} and in the second one the MLP model. All models are trained on a GTX-670MX of 3GB memory. Training the proposed architecture required 45 minutes.
\newline 
Figures \ref{one_min}, \ref{two_min} and \ref{three_min} depict the error rate for the ECML/PKDD-15 and Foursquare for 1, 2 and 3 minutes’ prediction.
Results demonstrate that DNN trained on GASF and GADF i.e. H-RNN and ResNet  outperform prediction models trained on the raw data i.e. MLP and IRNN.  For example, for Foursquare data and for 3 minutes ahead prediction, the proposed approach achieves an error rate of 0.025 while H-RNN leads to an error rate of 0.048.  With MLP, the achieved error rate is 0.1 while IRNN shows the worst performance with error rate of 0.16. Given the achieved error rates, the proposed architecture showed the best performance and is able to generalize quite well on the testing data with 47\% improvement compared to the closest performance.

\subsection{Discussion}
Our experiments showed that the DNN can exploit the data efficiently to achieve good prediction performance unlike classical machine learning algorithms. This confirms the fact that more data represent a burden for classic learning tasks. The abundance of the crowdsourced data represents a great opportunity to apply state-of-art deep learning techniques to make these data useful. Our aim is to benefit from these opportunities to provide a data-driven approach to enhance the crowdsourced service paradigm and tackle the problem of non-determinism of their spatiotemporal properties. However, one of the major challenges is to train these models on good representative data from crowdsourced sensors. Building such data is quite challenging, complex and time consuming and raises privacy issues. Furthermore, training very deep network such as ResNet-18 requires using adequate hardware. Fortunately, this task can be conducted offline using Graphics Processor Units (GPUs). Once trained, these models can be queried online to predict the service availability and its temporal attributes.

\section*{Conclusion}
We presented a deep learning based framework to predict crowdsourced service availability spatially and temporally. This framework is based on two  prediction models preceded by a pre-processing stage where we take into account the spatiotemporal features of the input data. In the first stage, we formulated the service discovery as a classification problem in which we associated every geolocation and time granularity to particular service providers. A DNN trained on these data can effectively predict the availability of services at location $l_i$ and time $t_i$. In the second stage, we formulated the prediction of the service availability duration as time series forecasting task. The time series are transformed into polar coordinates systems and the Gramian Angular Field are derived. This offered a new data representation of time series and enabled the application of  a deep learning-based forecasting technique.  A ResNet-based DNN trained on the generated images is able to predict the presence of the crowdsourced service for multi-step ahead .\newline
We conducted experimental studies to confirm the effectiveness of our framework. We have compared the service prediction using our deep learning-based approach with classical machine learning algorithms and assessed the performance of the proposed time series forecasting network against state-of-the-art deep learning based approaches. \newline
\section*{Acknowledgment}
This publication was made possible by NPRP grant \# NPRP9-224-1-049 from the Qatar National Research Fund (a member of
Qatar Foundation). The statements made herein are solely the
responsibility of the authors.

\end{document}